# Using an AI creativity system to explore how aesthetic experiences are processed along the brain's perceptual neural pathways

Vanessa Utz and Steve DiPaola

Simon Fraser University, Vancouver, Canada

`vutz@sfu.ca, sdipaola@sfu.ca`

**Abstract.** With the increased sophistication of AI techniques, the application of these systems has been expanding to ever newer fields. Increasingly, these systems are being used in modeling of human aesthetics and creativity, e.g. how humans create artworks and design products. Our lab has developed one such AI creativity deep learning system that can be used to create artworks in the form of images and videos. In this paper, we describe this system and its use in studying the human visual system and the formation of aesthetic experiences. Specifically, we show how time-based AI created media can be used to explore the nature of the dual-pathway neuro-architecture of the human visual system and how this relates to higher cognitive judgments such as aesthetic experiences that rely on these divergent information streams. We propose a theoretical framework for how the movement within percepts such as video clips, causes the engagement of reflexive attention and a subsequent focus on visual information that are primarily processed via the dorsal stream, thereby modulating aesthetic experiences that rely on information relayed via the ventral stream. We outline our recent study in support of our proposed framework, which serves as the first study that investigates the relationship between the two visual streams and aesthetic experiences.

## 1  Introduction

Over recent years, with the increased sophistication of AI systems and their expanded application, we have seen the more widespread use of these systems in fields related to the arts, aesthetics and creativity. The number of examples is extensive: when using Instagram, the "Explore" page provides users with image suggestions based on an analysis of images the user has previously engaged with, in the hopes of attracting the user to new content that they will like to a similar degree. In order instances, AI systems have gathered lots of media attention when they were used to create screenplays (see: Ars Technica, 2016) and original art pieces (see: Baraniuk, 2017). As the aim of these systems is to elicit aesthetic experiences that mimic the ones observers go through when encountering artworks created by human artists, these systems benefit from the integration of insights obtained from studying the human brain. Neuroscience, cognitive science and psychology have therefore become important sources for ideas and innovation regarding the design of such systems. Here however, we want to explore the reverse relationship and the use of these deep-learning systems as assistants in the knowledge building endeavor, to further the understanding of the human mind. We hope to be able to make use of creative AI systems to create artworks that mimic human-made artworks to such a degree that observers are unable to distinguish between the two. These artworks could then be tailored to researchers' specific needs to act as stimuli to study different aspects of human art perception. The focus of this exploration will be a newly developed framework regarding the dual-stream nature of the human visual system and reflexive attention, and how these play a role when observers are exposed to artworks involving motion, such as artistic videos. Our framework is based on the current understanding of aesthetics experiences and the formation of aesthetic judgments, as well as on knowledge regarding the human visual system. We therefore invest a section of this paper into outlining the current consensus in this research field, followed by a synthesis of this information into our framework. With this framework we aim to be provide the first exploration into how aesthetic experiences are affected by the way visual information are processed within the two streams of the visual system. We will begin the paper by providing an overview of the cognitively inspired deep learning creativity system that we are currently using, and which was used to create the stimuli for the study that was conducted to test our framework. Thereafter we will outline our framework and discuss our study which demonstrates initial evidence in support for our hypothesis.

## 2 Deep Learning Techniques applied in art, aesthetics and creativity research

When human artists create visual artworks, such as paintings, they engage in a variety of different cognitive mechanisms during their creative process. More often than not, paintings are not direct reflections of reality, but instead an expression of the artist's perception. Additionally, Zeki (2001) proposed that the motivation of artists could lie in the desire to reveal or trigger the neural underpinnings of visual perception. In order to better understand and even to potentially reproduce these creative processes, researchers have applied several different artificial intelligence techniques.

### 2.1 Application of cognitively inspired DL

Our lab has combined Genetic Algorithms, Neural Networks and Deep Learning neural networks into our Painterly Rendering software framework with the particular aim of emulating the cognitive processes of portrait artists (DiPaola et al., 2018, 2017, 2016, 2014, 2009). We will now review some of the work we have done in this field in order to demonstrate the use of biological inspired architectures in this space and to give the reader an idea of what kind of systems we are utilizing to create artworks that can be used for the study of human art perception.

### 2.2 Brief overview of the Deep Learning Techniques used

Deep Dream (Mordyintsev et al., 2015) and Neural Artistic Style (Gatys et al., 2015) are two techniques for modifying images through a process of analysis and search involving Deep Convolutional Neural Networks (DCNNs) (Krizhevsky, 2012; LeCun, 1998). We are abbreviating Deep Dream as DD and our modified Deep Dream system as mDD for the rest of the paper. DCNNs are typically trained on large datasets of images in order to build up a multi-level, feature-based re-encoding system, in which low-level features represent local features such as lines and curves, while high-level features represent more abstract, composite visual "concepts" such as "spoked wheel pattern" or "animal hind-leg shape". This method of representing images in a multi-layer network with increasing abstraction is thought to bear resemblance to the way the human brain processes visual perception (Di-Carlo, 2012). This structure facilitates performance in discrimination /classification tasks such as recognizing objects in an image as belonging to a certain learned category. However, as found with the DD systems, it is also possible to use DCNNs generatively, creating images which emphasize certain features or feature-layers of an image, or combine the features of one image with features from a second image to create an output image sharing qualities of both. These generative abilities resonate with a notion from the field of Neuroaesthetics (Zeki, 2001), namely that a possible role and motivation of art is for audiences/artists to reveal or stimulate the neural mechanisms of perception: we can view the different low- and high-level feature encodings within a DCNN as different perspectives on the essence of an image as analyzed within a human brain. In our previous work we have examined how DCNN and AI evolutionary systems can model different aspects of human cognition and creativity, including divergent thinking (DiPaola & Gabora, 2009), concept blending (McCaig, DiPaola & Gabora, 2016), and creative Honing Theory (DiPaola, McCaig & Gabora, 2018).

### 2.3 Our application of these systems

Within our AI-based painting software toolset, we have implemented our modified versions of DD using the Caffe deep learning framework (Jia, 2014). We are currently using them as a pre-processing stage which simulates an artists' imagination and perception, transforming an image before it is subsequently sent to the second, cognitive based, artistic stroke-placement phase (DiPaola, 2017). We now discuss the application of generative deep learning within our mDD system.

DD (Mordyintsev et al. 2015) has two basic modes of operation. The mode we call "free hallucination" begins with a source image and uses backpropagation and gradient ascent to gradually transform the image's pixels in order to emphasize the most strongly-activated features from a certain user-selected network layer. This process results in the emphasis of shapes and patterns that pre-exist in the source image, as well as the chance appearance of hallucinated patterns in which the network gravitates towards "seeing" patterns it has learned to recognize during its training on a data set. The other mode in which DD can be used, utilizes a biasing guide image. Again, through backpropagation and gradient ascent, the strong features from one "guide" image are analyzed and the best-matching features from a second source image are emphasized by transforming the image's pixels.

The DD algorithm emphasizing certain visual (semantic and/or stylistic) qualities, as illustrated in Figure 1. The algorithm begins by analyzing a human-selected guide image, which is propagated from the lowest (pixel) layer to a chosen higher layer. Higher layers encode the image in terms of progressively more abstract features. This encoding is stored as a guidefeature vector. Next, the algorithm initializes the set of pixels that constitute an updated source image which we refer to as the canvas image. The canvas image is gradually transformed into the output image. The canvas image is propagated up through the network to the same layer as the guide-feature vector, yielding a canvas-feature vector.

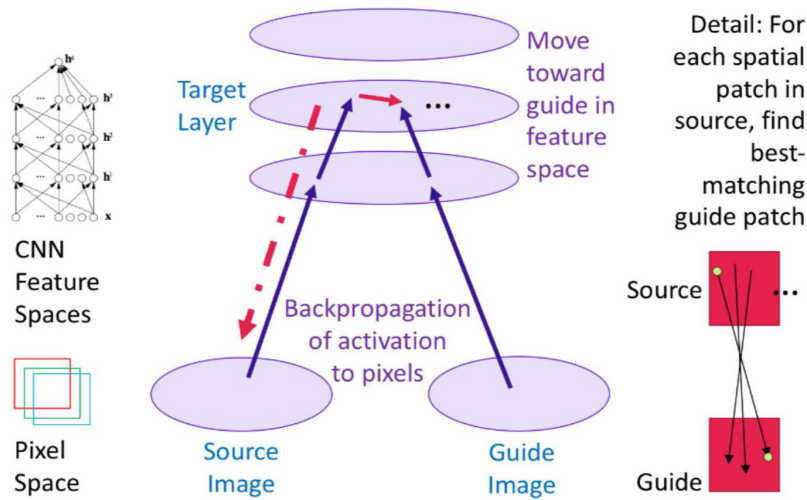

**Fig. 1.** Implementation and flow through of the Deep Dream Algorithm.

A loss function is defined to measure the difference between the two feature vectors. (The feature vectors are separated into spatial patches, and features from each spatial patch in the source are compared to the best-matching spatial patch in the guide. Thus, a dreamed output image tends to pick up on certain types of shape and texture, found within the guide image, that bear similarity to its own shapes and textures.) From this loss function, a gradient is found and back-propagated back to the lowest (pixel) layer. A small step in the direction thus found is applied as a change to pixel values. Repeating the cycle (of update pixels, propagate forward, find gradient, propagate downward) constitutes a gradient ascent process. Note that the original guided DD algorithm maximizes the dot product between source and guide features. This can result in textures and patterns that go beyond the guide features—the dotted line in Figure 1—into exaggerated or "hallucinatory" appearances. In contrast, for some of the work here, we have introduced variations on the algorithm, which instead minimize the distance between source and guide features. This tends to produce "tamer" output imagery with textures that preserve the look of the guide image.

To create a blend of two images, DD first generalizes each image by propagating it through a deep CNN and representing it according to the resulting tensor encoding at a certain network layer(s). Depending on the height and type of the network layer, the given encoding analyzes the image according to a particular set of visual features. Much as in the parable of the blind men who each describe an elephant in a different way, different layers "see the image" different ways. Thus, the visual blend depends on points of similarity between guide and source, as viewed from the perspective of a certain network layer encoding. In turn, the nature of a layer encoding depends on both the network architecture and the original training data which caused features to develop as a form of long-term memory. To enhance similarity between the source image and the guide, the algorithms use a reiterative two-phase creative process of alternating divergence and convergence: any similarities found at a high/abstract level are manifested back at the pixel level.

While most DD systems use pre-trained networks with object recognition data such as ImageNet, we have implemented our own modified DD system (mDD) and train new models with creativity art generation not object recognition in mind, using paintings and drawings as training data. We now have amassed a specific to fine art painting data set of 160,000 labeled / categorized paintings from 3000 labeled artists for a total size of 67 gigabytes of artistic visual data (one of the largest in an AI research group). Even with such a large and unique data set, during our AI training experiments, we discovered that since most fine artists make under 200 paintings in their lifetime (Picasso being an exception at 10 times that), we had material that might not be rigorous and large enough for an advanced CNN training for art styles. In order to overcome this issue, we developed a method which we call "hierarchical tight style and tile" (DiPaola et al., 2018). Since in our work, detecting and identifying regular objects 'within' an image is less important than the overall artistic style of the entire image (e.g. style of stroke, texture and color palette), we were able to use tile subsections of the images in order to increase our input data. We developed a hierarchical stochastic tiling method to sample each artwork into over 50 individual tiles of varying sizes. For instance, this allowed us to turn 50 source images of a category (e.g. Van Gogh's early work) into over 2000 individual training images for our network. This method allows us to train DD on the art style of fine art paintings instead of typical object recognition, providing superior output results. We also use a hierarchical and 'tight style' categorization labeling technique which is more valid for art texture and style than standard methods that typically categorize by artist name or a painting category. We have found that these types of labeled categories are too wide – meaning too visually diverse for good artistic style training. We therefore use a much tighter categorization method based on cognitive theories of art perception that we have outlined in previous work (DiPaola, 2014.). This method would sort Van Gogh's work into 7 hierarchical stylistic categories rather than one category. Figure 2 shows a source image (2a), which is processed through our mDD system (2b), then processed through our last step where it is rendered with cognitive based color and stroking in our ePainterly system for the final result (2c). This is the same process our study videos went through (Figure 4).

In this current system, we apply our mDD module to the source photo first, followed by our Non-Photorealisitc Rendering (NPR) painting phase. Our Painterly module, which we call ePainterly, is an extension to our cognitive painting system, Painterly (DiPaola 2009) and models the cognitive processes of artists based on years of research in this area. It uses algorithmic, particle system and noise modules to generate artistic color palettes, stroking and style techniques. It is the NPR subclass of stroke-based rendering that is used as the final part of our process to realize the internal DCNN models with stroke-based output informed by historic art making. Specifically, in this example, aesthetic advantages of this additional system include reducing noisy artefacting of the generated DCNN output via cohesive stroke-based clustering as well a better distributed color space (Figure 2c).

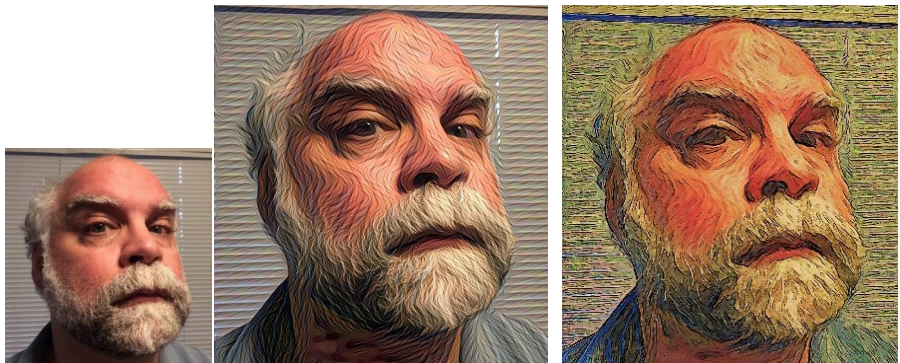

**Fig. 2.** a,b,c, Original source image (a, left), put through our modified Deep Dream systems (b, middle) then stroke and color enhanced through our ePainterly system for the final result (c, right).

## 3 Outline of the proposed cognitive framework

Since the aim is to use AI artworks to study human art perception, we will now spend some time discussing the instantiation of aesthetic experiences and the human visual system. What follows is a short summary of the two most commonly encountered models on aesthetic experiences, the dual-stream nature of the visual pathways and the engagement of attentional mechanisms, and how this information synthesizes into a testable hypothesis that can be investigated using AI art.

## 3.1 Existing models of aesthetic experiences

In the field of empirical aesthetics, which focuses on the study of aesthetic experiences in human observers, the most prominent model has been developed by Leder and colleagues in 2004 (Leder et al., 2004) with the aim of integrating the insights that had been obtained in psychological studies during the previous decades. Up until that point in time, research regarding aesthetic experiences had lacked a cohesive model that united all the different research foci. This cognitive model consists of a 5-step process, leading from a piece of visual artwork as input, to a two-fold output - an aesthetic judgment and an aesthetic emotion. The steps are as follows: (1) perception, (2) implicit memory integration (e.g. based on previous experiences – analysis of familiarity), (3) explicit classification (e.g. based on domain-specific expertise – content and style analysis), (4) cognitive mastering and evaluation and (5) continuous emotional evaluation which takes places throughout the entire process. This model has subsequently been slightly adapted in 2014 based on new research findings in this space, especially regarding the role of the emotions in the instantiation of aesthetic experiences (Leder & Nadal, 2014). This however is of less importance for our research since we are maintaining our focus on the perceptual aspects that play a role in the creation of aesthetic experiences.

Another commonly encountered model of aesthetic experiences was introduced by Chatterjee (Chatterjee & Vartanian, 2014). This so called 'Aesthetic Triad' illustrates the interaction between different cognitive processes that are involved in the instantiation of aesthetic experiences, based on neuroscientific evidence regarding the formation and nature of aesthetic judgements. The triad highlights the following systems: (1) emotion-valuation, (2) sensory-motor and (3) knowledge-meaning. Together these three components make up aesthetic experiences, however it has been pointed out that not all aesthetic experiences necessitate all three aspects (see Chatterjee & Vartanian, 2014). The Aesthetic Triad has helped to separate the individual mechanisms involved in aesthetic experiences and thereby allowed the mapping of these processes onto different aspects of the human neural architecture.

## 3.2 Perception, context and framing

Both Leder et al.'s (2004, 2014) model and the Aesthetic Triad (Chatterjee & Vartanian, 2014) emphasize similar aspects of aesthetic processing: perception, emotion and interpretation. Together they provide a comprehensive overview of the current state of knowledge and understanding regarding the multitude of processes involved, as well as function as foundations for the future study of this emerging field. Especially Leder et al.'s cognitive model has already been used in a multitude of studies to provide a foundational theoretical framework for the factors that are investigated in these studies (e.g. see: Belke et al., 2015; Gerger et al., 2014; Chamberlain et al., 2017). Influences that have been investigated in this field include a range of both top-down and button-up processes that are involved in the instantiation of aesthetic experiences. One of these variables is the framing of the artwork/the context in which the artwork is presented. It has been repeatedly found that an observer's perception of context and the artwork itself has a robust influence on the aesthetic experience (Gerger et al., 2014; Locher at al., 2015; Mullennix & Robinet, 2018; Nissel et al., 2015; Pelowski et al., 2017; Van Dongen et al., 2016). It has also been pointed out by Chatterjee & Vartanian (2014) that the systems involved in creating aesthetic experiences are also engaged when encountering non-art objects and that merely the framing of an object as artwork affects the interaction between the systems in such a way to allow for an aesthetic experience. This finding is of importance in our hypothesis regarding the perception and aesthetic evaluation of visually based art videos. For this work, we are defining 'art videos' as visual artistic work, similar to moving paintings, that have a movement and time based component to them. When encountering art-videos within a non-art context, we expect little change in how observers engage with them compared to any other regular video footage, due to the uncertain status of videos as art. We presume that a difference between how traditional paintings and art-videos are processed would only become apparent once observers are aware of their art status. If this is correct, then the presentation of our video stimuli as art is of utmost importance if we want to prevent context and the perception of our stimuli as non-art to act as confounding variables.

### 3.3 Processing visual information

Since our hypothesis is based on the different mechanisms that make up the visual system, we will briefly summarize the features of the visual system that are vital for our framework. The two aspects, which we are particularly going to focus on are the dual stream nature of the visual system and the mechanisms leading to reflexive visual attention.

**Dual-stream pathways in the visual system** The notion that our visual system consists of two separate pathways has become well established since the publication on the issue by Goodale and Milner (1992). The two pathways have become widely-known as the "what" and "where" pathways, referring to a ventral (occipito-temporal) and a dorsal (occipito-parietal) processing stream respectively (see Figure 3). The ventral stream relays information important for object recognition (such as color and form) to the Orbitofrontal Cortex (OFC) (Rolls, 2005). The OFC is a higher-level sensory cortex, part of the frontal lopes, and is of importance during decision making tasks. Rushworth and colleagues (2008) introduced a model that showed decision-making, which is involved in the appraisal of objects, is linked to activation in the OFC. Numerous neuroscientific studies investigating the appraisal of quality of both art and non-art, have shown that the OFC is activated during decision making tasks of this kind (see Kringelbach, 2005; Wallis, 2007). This connection between the ventral pathway and the OFC appears therefore to be of importance in the instantiation of aesthetic experiences and formation of aesthetic judgments (see Brown et al., 2011 for a cognitive model of the core aesthetic network in the human brain). On the other hand, the dorsal stream leads from the primary visual cortex to the parietal lobe and processes information important for the determination of location and movement of objects within the visual field.

**Visual attention** The other important aspect of the visual system is how visual attention is focused on the different parts of the visual scene that one is presented with. There appear to be competitive interactions between neural representations of the different visual stimuli that we perceive. Focusing attention on one of these stimuli enhances the neural response towards said stimulus (Desimone, 1998). Generally, behaviorally relevant stimuli are favored when it comes to attracting one's attentional focus, based on several top-down and bottom-up processes. One of these bottom-up processes involve abrupt changes that can occur within a visual field. The nature of a change can include a wide range of different characteristics, such as the disappearance or appearance of objects, or a color or shape change among others. These changes lead to reflexive attention, a process during which attention is rapidly oriented towards a change. Reflexive attention causes a modulation during the early stages of the sensory analysis and subsequently enhances higher level analyses at later stages. While these modulations engaged rapidly after the change in the visual system occurs, they are only short lived (Hopfinger & Mangun, 1998).

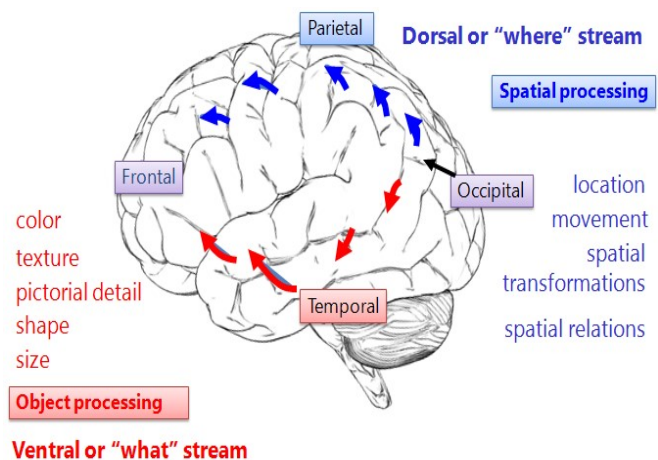

**Fig. 3.** Annotated diagram of the dorsal and ventral streams in the human visual system. Annotations highlight types of information that are processed via each stream (Kozhevnikov, 2019).

### 3.4 Synthesis into a single framework

We now demonstrate how the understanding of the creation of aesthetic experiences to visual artworks and the knowledge of the human visual system fit into one framework and how this has led to our hypothesis regarding the influence of movement perception on the formation of aesthetic judgments. Since the ventral visual pathway is mainly responsible for the processing and relaying of information regarding object recognition, such as color and detail, this pathway is of vital importance during the formation of aesthetic judgements (see Leder et al.'s (2004) model). In order for a piece of artwork to be appropriately evaluated, evidence regarding the content, artistic style, color combination and such are needed. The information that is required for this is primarily the type of information that is passed along the ventral pathway. Research has shown that when individuals are exposed to pieces of artwork, it takes approximately 3 seconds for the visual information to be adequately processed and for the person to produce an aesthetic judgement to the artwork. Shorter presentation times negatively affect an observer's ability to process all information necessary and thereby prevent adequate aesthetic judgments (Leder et al., 2004). If an observer however is presented with an art video, the continuous appearance of unexpected changes within the visual field are attended to. Reflexive attention is therefore likely to engage, and the dorsal stream of the visual system processes the continuously appearing and changing forms on screen. We therefore propose that due to the presence of competing visual information foci and the reflexive shift of attention towards changes in the visual field, aesthetic experiences in observers are altered when they encounter art videos, compared to what they would experience when observing a still image. More specifically, we expect that if tested they would provide lower ratings regarding the aesthetic liking and their aesthetic experience overall when presented with these art videos in comparison to more traditional art forms such as paintings. There has been extensive research into the types of information that the two visual streams process (see Figure 3), however this framework is the first attempt to determine the impact of the different processing streams on the aesthetic experience.

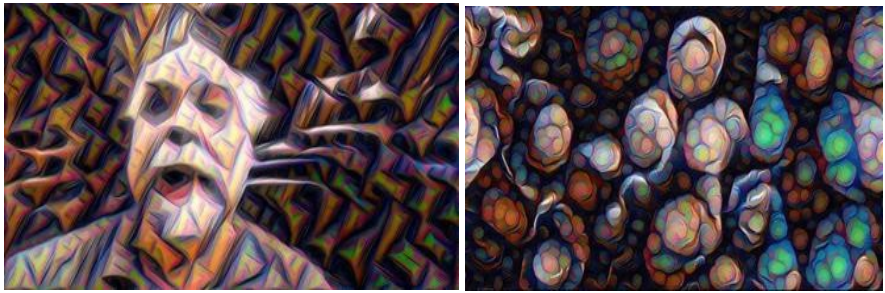

**Fig. 4.** a, b   Screenshots from our fast and slow movies - our 'portrait' video is a slightly less abstract art video (a, left). Our 'abstract' video shows 3 women in a highly abstract form (b, right).

## 4   Testing of the proposed framework

In order to test our framework, we have conducted an initial investigation into this subject matter. The following section will describe our methodology.

The aim of this study was to test our hypothesis and to gain some additional insights into how people perceive our stimuli. We therefore collected a combination of quantitative and qualitative data from our participants. The quantitative data was used to assess whether our framework could be validated, while the qualitative data (collected as answers to open ended questions) served to give more information that could potentially be used in future investigations.

### 4.1   Methods:

**Participants** Our participants were recruited at Simon Fraser University (SFU), Canada. The participant pool consisted of undergraduate and graduate students in SFU's School of Interactive Arts and Technology where they have some but limited knowledge of art. The undergraduate students received class credit for the participation. The graduate students were not compensated for their participation in any form. The age of the participants ranged from 18 years to 40 years. 14 were male, 24 were female. We recruited a total of 44 participants, however 6 had to be excluded from the data analysis, leaving us with N=38. Reasons for exclusion: 1 had a panic response to one of the video stimuli, 2 were shown longer versions (20s instead of the short 9s versions) of the slow stimuli videos, 3 did not understand the instructions or their responses on the answer sheets were ambiguous or incomplete. All participants were tested individually and all of them were naïve to the purpose and goal of the study. They were not told about our hypothesis and framework.

**Stimuli** We used two source videos of individuals that we turned into four art videos using a combination of mDD and the NPR Painterly module. We created the stimuli by first cutting the source footage into individual frames, and then using mDD to abstract each image. Each batch of frames was abstracted using a different 'recipe' from a list of network, guide image, layer, iteration, octave combinations that have previously been shown to produce quality output images. We ensured that each of the two batches of frames had been given a distinct artistic style (see Figure 4) for example frames of each of the two videos). Additionally, the videos were abstracted to different levels. Both original video footages contained humans, however for one video we abstracted the footage so that the human figures were no longer clearly visible ('abstract video'). In the other video, the human face remained clearly visible despite the addition of abstraction ('portrait video'). Using ffmpeg we then put the sets of frames back together into short videos ("How to speed up/slow down a video", 2019). We created two art videos from each set of frames, displaying the frames at different speeds, one in realtime (ffmpeg: setpts=1.0*pts for regular playback speed ('fast version')) and the other 3.5 times slower than realtime (ffmpeg: setpts=3.5*pts for the slow playback speed ('slow version')).

**Setup** We showed each participant two pairs of art videos. The videos were displayed in full screen mode on a laptop (a Lenovo Yoga 720 with 15.6" UHD display – 3840x2160 resolution) that was placed on a desk in front of the participant. Viewing distance was approximately 50cm. Order of video pairs (abstract vs portrait) and order of playback speed (fast vs slow), were randomized. To collect the quantitative measure, we asked the participants to rate each video on a 7-point Likert scale for A) liking (personal preference), B) aesthetic pleasantness (aesthetic beauty) and C) artistic value (how good of an art piece is the stimuli) (adapted from: Haertel & Carbon, 2014; Belke et al., 2015; Pelowski et al., 2018). Higher scores on these rating scales indicated higher agreement. Participants were asked to rate each video with respect to the other video in the pair. After providing their quantitative responses on a provided answer sheet, they were asked two open-ended questions: 1) "What was your experience watching the videos at different speeds?" and 2) "Which version of the video did you prefer? Explain why." Verbal answers were recorded and subsequently transcribed and analyzed. The purpose of these open-ended questions was to provide us with additional qualitative data regarding our participants' experiences while watching the different videos and to allow them to provide reasons for their preferences. Participants were allowed to speak freely for as long as they wanted. Experimenters only occasionally asked general questions in order to guide participants if their answers were off track or not detailed enough. Examples of such additional questions are for instance: "why did you think that?", "could you elaborate on this?", "since we are looking as these videos as artworks, could you add anything about your experience regarding more artistic concepts such as color and shape?".

## 4.2 Results and Discussion

The results we obtained in this study provide supporting evidence for our hypothesis that playback speed affects the way individuals form aesthetic experiences: the slow versions of the videos received higher scores on all three rating dimensions that we gave to our participants. On average, these slow versions were liked more and were judged to have higher aesthetic pleasantness and higher artistic values. Furthermore, 21 out of the 38 participants always preferred the slow versions for both video pairs (rated them with higher average score across the three rating dimensions). Only 3 preferred the faster versions both video pairs. The other 14 participants had differing preferences depending on which video was shown. 10 preferred the slow abstract video while preferring the faster version of the portrait video. For the other 4 participant this pattern was reversed (preference of slow portrait video and fast abstract video). This overall preference for the slower videos, was also supported by the data gathered from the answers to the open-ended questions.

**Quantitative Data** We obtained highly statistically significant differences in the scores for the two different speeds on all three rating dimensions. To compare the mean scores on these rating dimensions, we performed paired t-tests. For the likability scale, we obtained an average score of 5.46 (SD= 0.80) for the slowed videos and an average score of 4.46 (SD = 1.12) for the version played at regular speed, $t(37)=4.48$, $p<0.00001$ (On all three rating scales a value of '4' indicated the neutral midpoint). For the aesthetic pleasantness score, we obtained averages of 5.54 (SD=0.88) and 4.39 (SD=0.99) for the slow and fast versions respectively, $t(37)=5.35$, $p<0.00001$. Lastly, for the artistic value, our participants rated the slow videos with an average score of 5.37 (SD=0.93), while rating the faster versions with an average score of 4.83 (SD=0.95), $t(37)=2.50$, $p=0.01$. Since the average scores on each of the dimensions was significantly higher for the video versions with the slower playback speed, this data is supportive of our hypothesis.

Since the two video stimuli that we used were very different in nature (one was highly abstract with no recognizable individuals, while the other had a clear human face in the center of the frame) and based on some of the results we obtained from our qualitative data (discussed below), we also compared the effect of different playback speeds for the two videos separately. Our results indicated that the mean differences between scores were more significant for the highly abstract video. For the rating scales 'likability' and 'aesthetic pleasantness', the differences between group means (slow vs fast playback speed) were highly significant for both video types (likability: $t(37)=3.68$, $p<0.001$ for the abstract video, $t(37)=3.41$, $p=0.001$ for the portrait video; aesthetic pleasantness: $t(37)=5.52$, $p<0.0001$ for the abstract video, $t(37)=2.93$, $p<0.01$ for the portrait video). For the last rating dimension, artistic value, we only obtained a statistically significant difference for the abstract video, $t(37)=3.11$, $p<0.01$. For the portrait video, we obtained a smaller difference between scores: 5.32 (SD=1.01) for the slow version and 5.03 (SD=1.10) for the regular speed version, $t(37)=1.18$, $p=0.24$. This difference in scores was not statistically significant. This smaller difference is due to the higher average artistic value score that the portrait video received when played at regular speed, compared to the video's scores on the other two rating dimensions: 4.71 (SD=1.14) and 4.68 (SD=1.09), for likability and aesthetic pleasantness respectively.

**Qualitative Data** Participants spoke for an average of 82.4s (SD=31.7s). For the analysis of the recorded data, we transcribed the audio files and looked through the most representative responses, as well as outliers. Importantly, all participants commented on a noticeable difference in their perception of the videos based on the playback speed (example quote of participant: "It was super interesting to see how much the speed actually changed the art piece. It changed the way I interpreted the art piece […]"). Other quotes will be presented and discussed below.

As evidenced by the quantitative data, participants generally preferred the slow versions of the videos. In these instances, participants regularly commented on the fact that the slower playback speed allowed them to observe the finer details, and color and pattern changes: "[…] I was able to […] appreciate what was going on, see all the different little strokes. It was definitely a lot more pleasant.", "[…] in the slow videos I was able to see the blending of the colors […]", "[…] the colors seemed to stand out a lot more and I saw the definitions of a lot more lines included", "[…] you can see the patterns more clearly […]", "the environment is more clear. Clear to see the whole video". Additionally, participants repeatedly referred to the slower versions as more natural, calming and relaxing: "[…] it feels more pleasant, because it feels more normal […]", "[…] I could really relax and focus on what was going on […]", "[…] the slower version had a much more calming effect, where the movement just seemed a lot more soothing.", and that the movements appeared smoother. Some participants provided more interesting observations such as that their gaze felt to be less distracted: "If it's slowed down, my eye won't jump around those strokes". Particularly this last observation fits our hypothesis that the slower movement that is present in the slower versions of the videos, engages less attentional resources that could distract from the processing of information required for the formation of aesthetic experiences. In comparison, these participants that preferred the slower videos, perceived the faster versions as bumpy ("[…] in the fast one it was bumpy. That's why I didn't like the fast one."). This fits with other observations that participants reported to have a harder time focusing on smaller details and changes: "[…] the experience are different depending on the speed of the video, because if the videos with those small details are moving too fast, it is hard to capture how it changes because it disappears suddenly […]", "at the faster speed […] it was harder to tell what was going on […].", "[…] it was just a lot to register […]". One participant summarized their experience as such: "there was sort of a pulsating motion […] when it was fast. But when it was slow that all went away, and you just had a sense of all the colors […]". It was also commented on the impression that the fast version appeared less art-like and more like a natural video: "the faster one just felt like a video".

There were however participants that preferred the faster versions to the slowed videos. These participants highlighted that they perceived the colors in the faster videos to be more vibrant and saturated ("[…] they seemed more vibrant […]", "it seemed more vibrant in color, in the hues, yeah more saturated […]"), and that the lines appeared more defined: "the sped-up versions seemed more defined". There were also mentioning of a general preference of the faster versions due to their more natural flow of movement compared to the slowed version: "I like the flow. It just flows better", "[…] it seems more natural in movement […]", "[…] with the human face, when it is slowed down, it looks a little bit weird".

Just like in our quantitative data, in the qualitative data set we also observed evidence that the portrait video appeared to be less effective than the abstract video set in eliciting a response towards the different playback speeds. This is exemplified with the following quotes: "[…] because of the movement of the head, I couldn't figure out much of a difference. I could understand there was a speed variance, but from an art perspective I didn't see much of a difference." And "I thought both gave the same artistic value.".

Two of our participants made the interesting observation that the order of presentation might have affected their experiences: "When I watched the fast one first and then watched the slow one, it affected me because it felt glitchy, laggy, like a video that is buffering and slows down. That kind of put me of. But for the second set it was different because you showed me the slow one first and that kind of changed things. When you showed me the slow one first, I felt the fast one was too fast.".

## 5      Future Work and Conclusion

In this paper we introduced our new framework regarding the interaction of the visual system's dual pathways and attentional system with the formation of aesthetic experiences when exposed to moving art pieces. This paper therefore serves as the first investigation of the relationship between the visual pathways and aesthetic processing. While the difference in processing of color and detail between the visual pathways has been established, this is the first exploration of the impact on aesthetic experiences. We demonstrated that our framework is founded on existing and validated knowledge regarding the nature of the human visual system and how aesthetic experiences are formed within individuals. Through our initial study into this new framework, we have been able to demonstrate that there is supporting evidence for our proposal that the presence of movement within a visual percept interferes with the formation of aesthetic experiences due to the engagement of reflexive attention and the dorsal visual pathway. Our hypothesis was that if participants are presented with art videos, their aesthetic judgment of the art pieces would be higher for the versions with slower playback speed. Both our quantitative and qualitative data sets supported this hypothesis and showed that participants provided higher scores for the slower versions on all rating dimensions and commented extensively on how they were able to better perceive detail and color. Our current work therefore serves as a foundation for future work. Despite this initial data that we were able to successfully obtain, to fully validate our framework, more studies in this area need to be conducted. The next study needs to be of a major scale in order to fully investigate this multivariate phenomenon. It needs to provide rigorous data sets to support our claims and to further explore the new insights that we have gained with this initial investigation. We will outline some of these considerations.

The most important suggestion for future studies regards the stimuli. Our stimuli set size in this study was very limited. We only used two videos which we played at different speeds. In order to show the universality of the effect, future studies should use larger stimuli sets. Additionally, the difference between the effect in the abstract versus the portrait video needs further investigation. This is of importance, since both data sets indicated a difference in the way our participants reacted to the two videos. It is also worthwhile investigating whether the presence of clearly visible human faces or just the lower level of abstraction caused the lower effect size for the portrait videos. If the clearly visible human face is the cause of this, it should be investigated if the expectation of how fast a human usually moves their facial features affects the results, versus for instance the face acting as a distraction from the aesthetic evaluation of the rest of the artwork.

Since some of our qualitative data appears to be in conflict with each other, for instance with some participants perceiving the slow versions as more "natural" while others perceived the faster versions as more "natural", future studies should take a closer look at the possibility that individuals have a preferred speed at which they perceive videos and how this interacts with their notion of art and aesthetics.

Once other behavioral studies show similar evidence for our framework, it would be necessary to involve brain imaging techniques in subsequent research. These would allow insights into the different brain areas that are used when watching art videos compared to more traditional artworks such as paintings. Another avenue in this area would be lesion studies investigation whether individuals suffering from lesions in brain regions involved in the dorsal visual pathway, provide different aesthetic judgments compared to healthy individuals. These additional studies would serve to validate our proposed cognitive mechanism and would aid in the expansion of our methodology (using AI systems in the creation of tailored visual stimuli) into the field of neuroscience.

## Acknowledgements

We acknowledge our colleague Graeme McCaig who was instrumental in creating our modified Deep Dream system and was invaluable in his mentorship of the work. This work was partially supported by SSHRC and NSERC grants respectively.

17